\RequirePackage{amsthm}
\documentclass[pdflatex,iicol,sn-mathphys-num]{sn-jnl}

\usepackage{lineno}  
\usepackage{anyfontsize}
\usepackage{lmodern}  
\usepackage{graphicx}%
\usepackage{diagbox}
\usepackage{multirow}%
\usepackage{amsmath,amssymb,amsfonts}%
\usepackage{amsthm}%
\usepackage{mathrsfs}%
\usepackage[title]{appendix}%
\usepackage{xcolor}%
\usepackage{textcomp}%
\usepackage{manyfoot}%
\usepackage{booktabs}%
\usepackage{algorithm}%
\usepackage{algorithmicx}%
\usepackage{algpseudocode}%
\usepackage{listings}%
\usepackage{hyperref}

\theoremstyle{thmstyleone}%
\theoremstyle{thmstyletwo}%
\theoremstyle{thmstylethree}%

\begin{document}

\title[Article Title]{Enhanced Multi-Tuple Extraction for Alloys: Integrating Pointer Networks and Augmented Attention}

\author[1]{\fnm{Mengzhe} \sur{Hei}}\email{heimengzhe18@nudt.edu.cn}
\equalcont{These authors contributed equally to this work.}
\author*[2]{\fnm{Zhouran} \sur{Zhang}}\email{zzhang\_nudt@outlook.com}
\equalcont{These authors contributed equally to this work.}
\author[3]{\fnm{Qingbao} \sur{Liu}}\email{liuqingbao@nudt.edu.cn}
\author[1]{\fnm{Yan} \sur{Pan}}\email{panyan@nudt.edu.cn}
\author[3]{\fnm{Xiang} \sur{Zhao}}\email{xiangzhao@nudt.edu.cn}
\author[2]{\fnm{Yongqian} \sur{Peng}}\email{pengyqnudt@nudt.edu.cn}
\author[2]{\fnm{Yicong} \sur{Ye}}\email{18505993519@163.com}
\author*[1]{\fnm{Xin} \sur{Zhang}}\email{shinezhang\_nudt@163.com}
\author[2]{\fnm{Shuxin} \sur{Bai}}\email{shuxinbai@hotmail.com}

\affil*[1]{\orgdiv{National Key Laboratory of Information Systems Engineering}, \orgname{National University of Defense Technology}, \orgaddress{\city{Changsha}, \postcode{410072}, \state{Hunan}, \country{China}}}

\affil[2]{\orgdiv{Department of Materials Science and Engineering}, \orgname{National University of Defense Technology}, \orgaddress{\city{Changsha}, \postcode{410072}, \state{Hunan}, \country{China}}}

\affil[3]{\orgdiv{Laboratory for Big Data and Decision}, \orgname{National University of Defense Technology}, \orgaddress{\city{Changsha}, \postcode{410072}, \state{Hunan}, \country{China}}}

\abstract{
Extracting high-quality structured information from scientific literature is crucial for advancing material design through data-driven methods. Despite the considerable research in natural language processing for dataset extraction, effective approaches for multi-tuple extraction in scientific literature remain scarce due to the complex interrelations of tuples and contextual ambiguities. In the study, we  illustrate the multi-tuple extraction of mechanical properties from multi-principal-element alloys and presents a novel framework that combines an entity extraction model based on MatSciBERT with pointer networks and an allocation model utilizing inter- and intra-entity attention. Our rigorous experiments on tuple extraction demonstrate impressive F1 scores of 0.963, 0.947, 0.848, and 0.753 across datasets with 1, 2, 3, and 4 tuples, confirming the effectiveness of the model. Furthermore, an F1 score of 0.854 was achieved on a randomly curated dataset. These results highlight the model’s capacity to deliver precise and structured information, offering a robust alternative to large language models and equipping researchers with essential data for fostering data-driven innovations.
}




\maketitle
\linenumbers 
\nolinenumbers
\section{Introduction}
\label{sec1}
The traditional trial-and-error paradigm in the development of new materials is protracted and exorbitant. However, with the advent of emerging technologies such as artificial intelligence, a dual-driven scientific artificial intelligence strategy—leveraging both models and data—has achieved widespread utilization in the design and development of novel materials\cite{gormley2021machine,wang2024accelerating}, as well as in elucidating the interrelationships between structure and performance\cite{carvalho2022artificial, bhowmik2022implications,zhou2023generative}. This paradigm has unveiled considerable potential for the efficacious design and optimization of materials\cite{basu2022biomaterialomics,singh2020artificial,Debnath2021GenerativeDL,hart2021machine}. Undoubtedly, data remains the foundational and critical element in achieving the aforementioned objectives. However, for certain material properties—particularly those pertaining to performance under service conditions—available data are still scarce. The acquisition of high-quality data incurs significant costs, rendering it difficult to meet the requirements for training models that demand high accuracy and robust performance.
\par
Scientific literature encompasses a vast array of peer-reviewed, high-quality, and relatively reliable data, serving as a crucial resource. However, this information predominantly exists in the form of unstructured text within a diverse array of sources, such as textbooks, material handbooks, patents, and research articles, rendering it unsuitable for direct use as structured data. Techniques for the automatic extraction of data and information regarding materials consequently present a promising opportunity to construct extensive databases for machine learning and data-driven methodologies. 
\par
Natural Language Processing (NLP) is an interdisciplinary field that integrates linguistics and artificial intelligence. One of the most prevalent applications of NLP technology is the automated extraction of structured information, facilitating the efficient mining of data from semi-structured tables and unstructured texts\cite{kononova2021}. Numerous studies have reported various methodologies for mining structured information within specific domains of materials science\cite{Sierepeklis2022,Kumar2022}, with a general trend moving from generic approaches to specialized techniques. Initially, this process relied on traditional processing pipelines that generally include named entity recognition (NER) and relation extraction\cite{wang2022}, such as ChemDataExtractor\cite{swain2016} for chemical information extraction. The specific techniques involved in the workflow include look-ups, rule-based\cite{wang2022}, machine learning methods which gradually shifted towards adopting more sophisticated deep neural models\cite{widiastuti2019}, such as Long Short-Term Memory (LSTM) networks\cite{wang2022}. Due to the excellent performance of pre-training and fine-tuning paradigm, many works relied on pre-trained models for extraction emerge. Through pre-training on a substantial corpus of materials-related literature, this model enables a more profound analysis of the relationships among chemical compositions, structural features, and their corresponding properties. Many variations of BERT\cite{Devlin2019BERTPO}, such as MatSciBERT\cite{Gupta2021MatSciBERTAM}, MaterialsBERT\cite{shetty2023general}, BioBERT\cite{lee2020biobert}, and BatteryBERT\cite{Huang2022BatteryBERTAP} have shown improved performance on downstream tasks. Large language models such as GPT-3/4 \cite{Brown2020LanguageMA,Achiam2023GPT4TR}, LLaMA\cite{Touvron2023LLaMAOA}, Palm \cite{chowdhery2023palm}, Gemini\cite{team2023gemini} and FLAN\cite{weifinetuned} have shown great understanding of textual information. With designed prompt, large language models can extract demanded information and data from literature. Dagdelen et al.\cite{Dagdelen2024StructuredIE} used multiple large language models to extract structured information. 
\par
However, despite the extensive research conducted on the extraction of structured information, effective methodologies for the challenge of multi-tuple extraction remain conspicuously lacking. A tuple typically consists of a set of related elements that represent structured information, often in the form of (subject, predicate, object) or (entity, attribute, value), although in some cases, the number of elements in a tuple can exceed 3. Due to the presence of multiple tuples within the text which leads to challenges in accurately assigning extracted entities to their corresponding tuples, coupled with the ambiguity arising from contextual dependencies and the diverse modes of expression, as well as the difficulty in obtaining high-quality annotated datasets, multi-tuple extraction continues to pose a formidable challenge. Even though large language models have demonstrated outstanding capabilities in natural language processing tasks, they still exhibit limitations in understanding multi-tuple scientific texts and may produce hallucinations. Moreover, it is noteworthy that large language models typically require substantial computational resources for both training and inference, which may pose challenges in terms of cost and efficiency in practical applications of data mining and knowledge extraction.
 \par
In this study, we explored the multi-tuple extraction of mechanical properties from multi-principal-element alloys (MPEAs) and proposed a novel approach that integrated an entity extraction model leveraging MatSciBERT and pointer networks, alongside an entity allocation model that employs inter-entity and intra-entity attention mechanisms. We constructed a corpus consisting of 255 sentences, each containing a varying number of tuples derived from scholarly literature on the mechanical properties of multi-principal-element alloys. This corpus encompasses a total of 568 tuples, each complete tuple comprising the alloy name, mechanical property, property values, conditions, and corresponding condition values. The literature often omits the last two entities when the measurement conditions are room temperature; however, the first three entities are never omitted in the dataset. To rigorously evaluate the model's performance across datasets with disparate tuple counts (1, 2, 3, and 4) within individual sentences, we partitioned the corpus into four distinct datasets based on the number of tuples and conducted evaluations for each. Sentences containing more than five tuples were excluded due to their insubstantial prevalence. Additionally, we randomly sampled a subset of sentences from the corpus to create a test set, thereby enabling us to gauge the model’s average performance comprehensively. Crucially, the evaluation of the model's efficacy is predicated upon the accuracy and completeness of the generated tuples rather than individual entities. The F1 scores for the proposed model were 0.963, 0.947, 0.848, and 0.753 for datasets containing 1, 2, 3, and 4 tuples, respectively, while the model achieved an F1 score of 0.854 on the randomly selected dataset. Moreover, the proposed approach outperformed four distinguished large language models (Claude 3 Haiku, GPT-4o, Gemini 1.5, and Llama 3.1 70B) within a prompt-answering framework.
\section{Results}
\subsection{Multi-tuple analyses}
\begin{figure}[H]
	\centering
\includegraphics[width=1\linewidth]{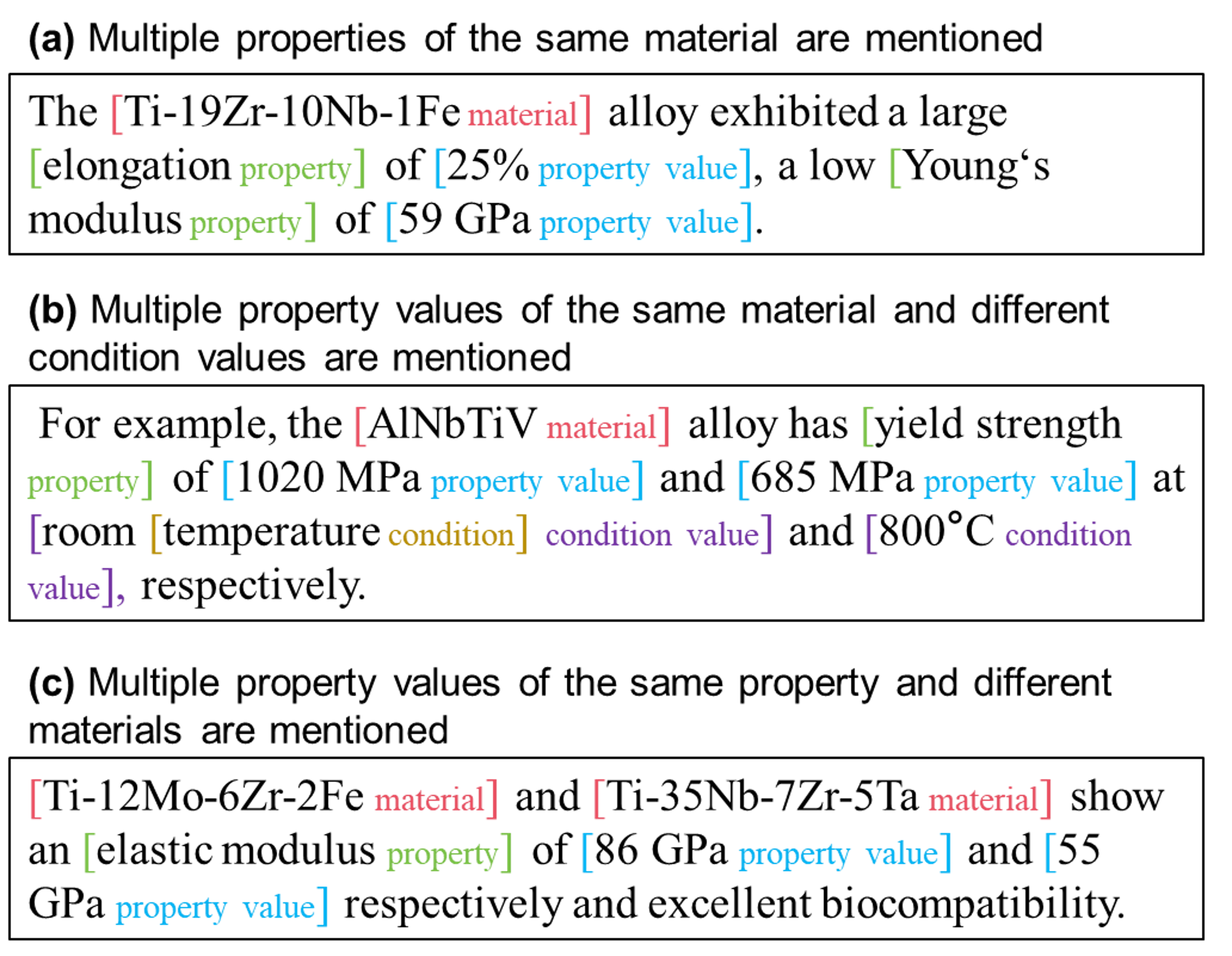}
	\caption{\textbf{Common patterns of multiple entities and multiple relations in multi-principal-element alloys}.  We use three simple sentences to exemplify three common repetition patterns. However, in real-world scenarios and our dataset, the input text comprises multiple sentences.
    \textbf{a} An example of multiple properties of the same material. \textbf{b} An example of multiple property values of the same material and different condition values. \textbf{c} An example of multiple property values of the same property and different materials.}
	\label{fig:types of multi}
\end{figure}

We delineate several prevalent multi-tuple patterns observed in the aforementioned situations, as illustrated in Fig. \ref{fig:types of multi}. These patterns include multiple properties of the same material, various property values for the same material under different conditions, and multiple property values for the same property across different materials. Furthermore, these common patterns can be interwoven within more intricate textual contexts, which are frequently encountered in the literature pertaining to diverse materials. For instance, from the sentence depicted in Fig. \ref{fig:types of multi}(\textbf{b}), we can extract two tuples: ["AlNbTiV", "yield strength", "1020 MPa", "temperature","room temperature"] and ["AlNbTiV", "yield strength", "685 MPa", "temperature","800°C"]. In this context, the mere extraction of individual entities proves inadequate for conveying the scientific significance embedded within the literature.

\begin{figure}[H]
	\centering
	\includegraphics[width=1\linewidth]{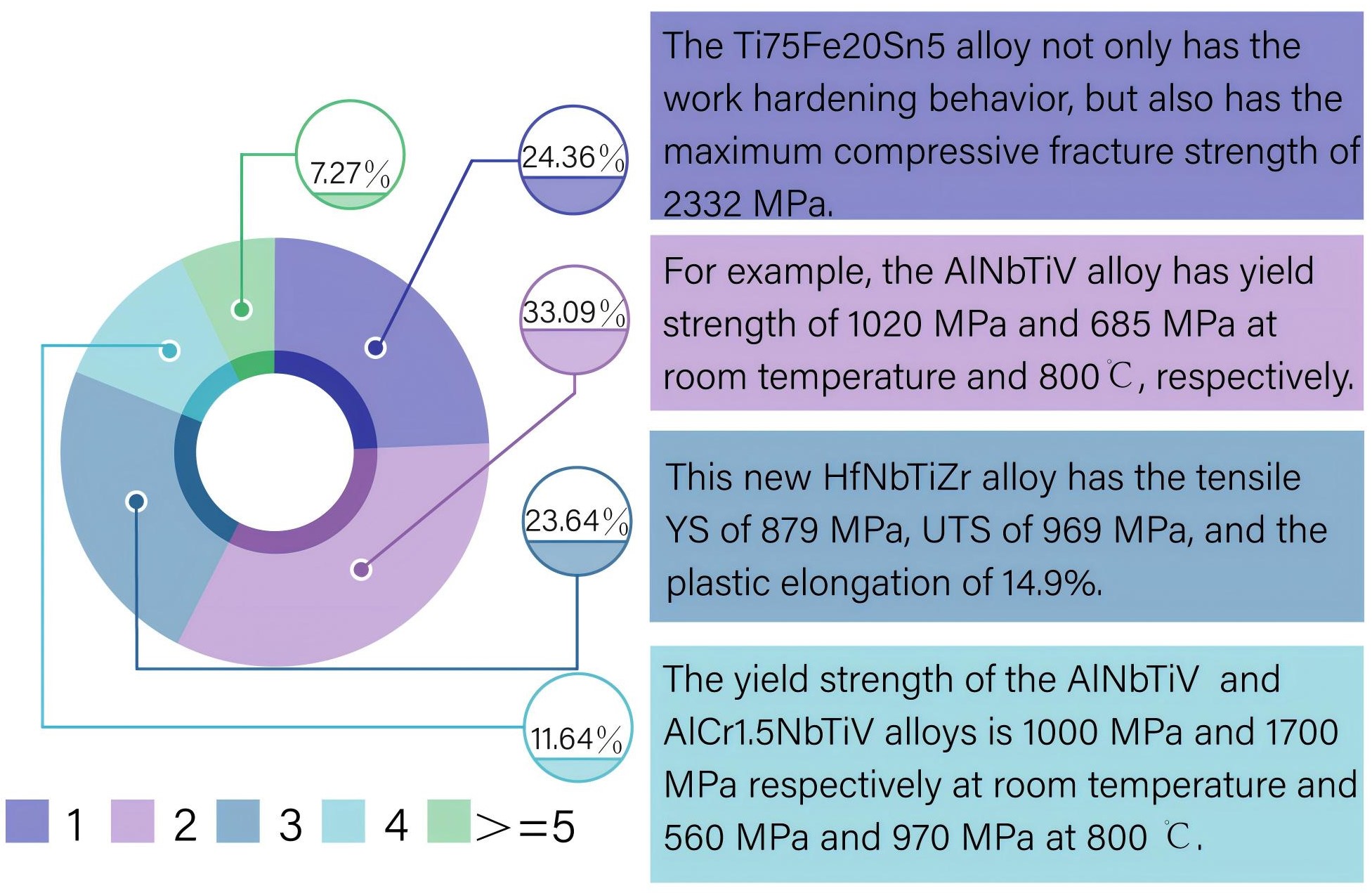}
	\caption{\textbf{The proportion of sentences containing different numbers of tuples.} The numbers below the pie chart indicate the number of tuples contained in each colored segment of the pie chart. On the right side of the pie chart are examples of varying number of tuples within one sentence. The proportions, in ascending order, are 24.36\%, 33.09\%, 23.64\%, 11.64\%, and 7.27\%.}
	\label{fig:account}
\end{figure}

Remarkably, it is the complete \textbf{tuples}, consisting of various types of entities, that must be extracted. Moreover, the presence of multiple tuples conveying scientific meaning within a single sentence is prevalent in the materials science literature. Our statistical analysis, derived from over a hundred publications on MPEAs, reveals that sentences describing alloy properties containing only a single tuple account for merely 24.36\%. The distribution of sentences containing varying numbers of tuples is illustrated in Fig. \ref{fig:account}. Nevertheless, there remains a paucity of studies that have proposed effective and sophisticated solutions to rectify this issue.

\begin{figure*}[htbp]
	\centering
	\includegraphics[width=\linewidth]{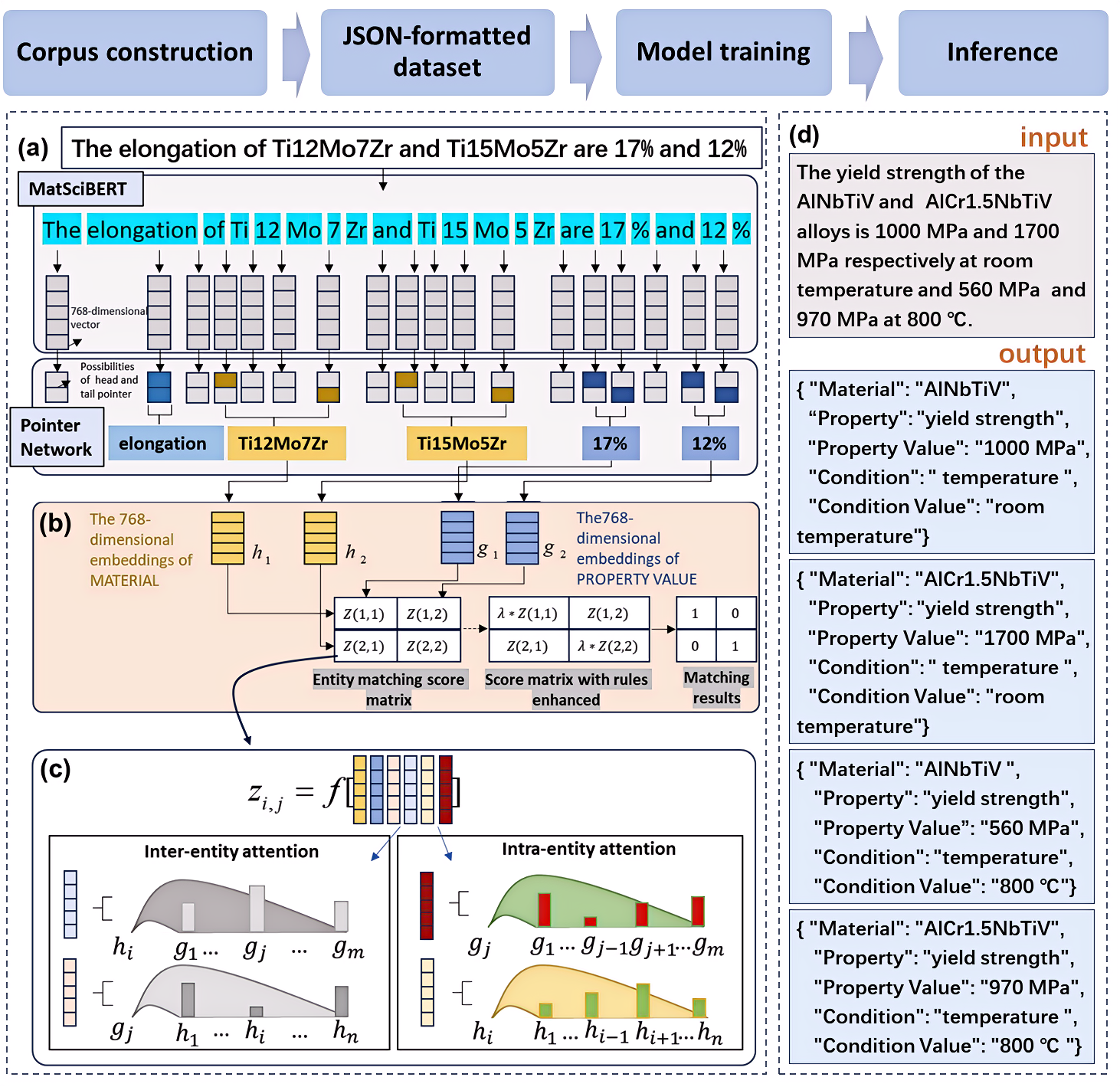}
	\caption{\textbf{The workflow and the framework of the proposed model for extracting and allocating entities.} 
		The workflow is presented in the upper section of the figure, beginning with the retrieval of full-text research articles from Elsevier, followed by the construction of a specialized corpus, from which we extract and annotate sentences to obtain a JSON-formatted dataset, and ends with model training and inference. The model is primarily composed of two components: entity extraction and entity allocation. \textbf{a} \textbf{Entity Extraction}: This component integrates MatSciBERT and a pointer network. MatSciBERT first tokenizes the input sentence and generates vector representations for each token. The pointer network then computes the probability of each token serving as the head or tail token of a specific entity, thereby identifying entities based on these probabilities. \textbf{b} \textbf{Entity Allocation}: This component assesses whether entities of different types belong to the same tuple through an entity matching score matrix. During model inference, we can enhance the matching likelihood of entities in corresponding order by multiplying the diagonal elements of the matrix by a parameter.  \textbf{c} \textbf{Entity Matching Score Matrix}: Each element in the matrix represents a combination of six vectors. The first two vectors correspond to the vector representations of the two entities, while the remaining four vectors are derived from two attention mechanisms: intra-entity and inter-entity attention. Intra-entity attention focuses on the attention distribution among different types of entities, whereas inter-entity attention concentrates on the attention distribution within the same type of entity. \textbf{d} \textbf{Inference}: A four tuple extraction example.}
	\label{fig:framework}  
\end{figure*}

Fig. \ref{fig:framework} presents the overall workflow of the study. Initially, we constructed a corpus comprising over 200 full-text papers containing the keywords "multi-principal-element alloy" and "mechanical properties." Subsequently, we extracted sentences from this corpus that describe alloy properties and annotated five distinct types of entities which are MATERIAL, PROPERTY, PROPERTY VALUE, CONDITION, CONDITION VALUE within these sentences. This process yielded 255 sentences containing varying numbers of tuples, resulting in a total of 568 tuples stored in a JSON-formatted file. Our model operates in two stages: entity extraction and entity allocation. First, given the diverse ways entities are expressed in the literature and the excellent performance of pre-trained models, we utilize a method based on the MatSciBERT model and pointer network\cite{sun2018multi,vinyals2015pointer} for entity extraction from input texts. MatSciBERT model\cite{Gupta2021MatSciBERTAM}, a BERT-based model specifically pre-trained on a substantial corpus of materials science literature that outperforms the BERT and SciBERT model in a series of downstream tasks in the material field, including entity extraction, is likely to bring about superior textual embeddings. It is then aimed to address the challenges faced by task-agnostic models in comprehending specialized terminology within the materials science domain by utilizing an embedding layer based on MatSciBERT. Unlike BERT, MatSciBERT dynamically generates masks in word level when providing textual input to the model during pre-training instead of during preprocessing, and MatSciBERT removes Next Sentence Prediction task. More importantly, MatSciBERT is pre-trained on full-length sequences of text, enabling it to handle longer sequences and cross sentence dependencies. This is crucial because the multi-tuple extraction problem addressed in this paper frequently involves long textual inputs.
\par
The pointer network is an extraction paradigm that trains two binary classifiers to predict the start and end tokens of an entity within the input text, thereby extracting the entity by identifying its start and end. This allows model to extract entities of variable length and position, such as the chemical formulas of newly synthesized materials, which always begin with a chemical element name and end with either a chemical element name or a number. Pointer network can also address nested entity problems in materials science. For example, in the text fragment 'room temperature' which is shown in the Fig. \ref{fig:types of multi} \textbf{b}, the CONDITION NAME 'temperature' and the CONDITION VALUE 'room temperature' are nested entities. Furthermore, pointer networks generate entity positions directly from the vector representations of token during decoding, eliminating the need for additional processing and increasing efficiency. This extraction paradigm performs better on well-formatted text and is more efficient than other methods due to its smaller search space, making it well-suited for information extraction scenarios such as entity recognition, relation extraction, and event extraction. 
\par
In order to tackle the challenge of multi-tuple extraction, we propose the task of entity allocation to assign entities of different types extracted by entity extraction to a complete tuple and avoiding tuple allocation errors caused by entity relationship confusion. We believe that the key to this task is to enable the model to learn both correct and incorrect tuple matching patterns contrastively. Hence, in the second stage, we construct an entity matching score matrix to assess the likelihood of various entities being allocated together. We apply inter-entity and intra-entity attention mechanisms to generate vector embeddings rich in semantic information. Specifically, the former is responsible for generating attention representations between different types of entities, while the latter focuses on generating attention representations between entities of the same type. This design allows the model to utilize information from the two entities being matched and all entities of the two types when performing tuple matching. This step is crucial for determining whether the output tuples accurately convey the intended scientific meaning, as any erroneous entity allocation by the model could distort the semantic content of the tuples. It is important to note that we do not directly predict the relationships between captured entities; rather, we ascertain whether two entities should reside within the same tuple. This approach is justified for two reasons: first, the number of relationships between entities in materials literature is fixed and limited once the types of entities to be extracted are identified, rendering relationship prediction less beneficial for extracting correct tuples. Second, predicting relationships between entities constitutes a multi-class classification task, whereas entity matching is a binary classification task, making the latter less prone to errors. In this stage, the five types of entities mentioned previously will be extracted for subsequent assignment to the correct tuples. 

Throughout this study, we utilized five test datasets. The differentiation among the first four test sets lies in the number of tuples encompassed within their respective sentences, specifically 1, 2, 3, and 4 tuples. Sentences containing more than four tuples were excluded due to their scarcity, rendering them impractical for training purposes. The fifth dataset was generated from the complete dataset through a random sampling procedure. Detailed information regarding the test sets is presented in Table \ref{tab:test data} (these datasets are denoted as 1, 2, 3, 4, and Random in the table, and this notation is consistent throughout the subsequent sections). As for the entire dataset, the specific information are delineated in the Table \ref{tab:total dataset}.
\begin{table}[h]
	\caption{Number of sentences and tuples of the test sets}\label{tab:test data}%
	\begin{tabular}{@{}llllll@{}}
		\toprule
		Test dataset & 1  & 2 & 3 & 4 & Random\\
		\midrule
		Num of Sentences    & 40   & 38  & 38 & 22 & 23 \\
		Num of tuples  & 40   & 76  & 114 & 88 & 50    \\
		\botrule
	\end{tabular}
\end{table}

\begin{table}[h]
	\caption{Number of sentences and tuples of the entire dataset}\label{tab:total dataset}%
	\begin{tabular}{@{}llllll@{}}
		\toprule
		dataset & 1  & 2 & 3 & 4 & total\\
		\midrule
		Num of Sentences    & 67   & 91  & 65 & 32 & 255 \\
		Num of tuples  & 67   & 182  & 195 & 128 & 568    \\
		\botrule
	\end{tabular}
\end{table}
\par
We conduct five training sessions, each time excising the respective test set from the entire dataset, utilizing the residual data as both the training and validating sets in a ratio of 9:1. 

\subsection{Performances of Entity Extraction}
When evaluating the model’s performance on entity extraction, we show the results on different types of entities, namely, MATERIAL, PROPERTY, PROPERTY VALUE, CONDITION, and CONDITION VALUE. We use the common F1, Precision, and Recall metrics, and the specific results are shown in the Table \ref{tab:extraction}. 
The metrics are calculated in the following manner:
\begin{equation}
	\begin{aligned}
		\text{Precision} & = \frac{\text{Number of correct entities extracted }}{\text{Number of entities extracted} } \\
		\text{Recall} & = \frac{\text{Number of correct entities extracted }}{\text{Number of entities in test set }} \\
		\text{F1} & = \frac{2 * \text{Precision} * \text{Recall}}{\text{Precision} + \text{Recall}}
	\end{aligned}
\end{equation}
\begin{table*}[h]
	\caption{Performances of proposed model on entity extraction}\label{tab:extraction}
	\setlength{\tabcolsep}{0.5mm}{
		\begin{tabular*}{\textwidth}{@{\extracolsep\fill}l|ccc|ccc|ccc|ccc|ccc}
			\toprule%
			& \multicolumn{3}{@{}c@{}}{1} & \multicolumn{3}{@{}c@{}}{2} &
			\multicolumn{3}{@{}c@{}}{3} &
			\multicolumn{3}{@{}c@{}}{4} &
			\multicolumn{3}{@{}c@{}}{Random} 
			\\\cmidrule{2-4}\cmidrule{5-16}%
			Entity & \textbf{F1} &\textbf{P} &\textbf{R} & \textbf{F1} &\textbf{P} &\textbf{R} &\textbf{F1} &\textbf{P} &\textbf{R}& \textbf{F1} &\textbf{P} &\textbf{R} & \textbf{F1} &\textbf{P} &\textbf{R} \\
			\midrule
			MATERIAL  & 0.963 & 0.951 & 0.975 & 0.97 & 0.942 & 1 & 0.941 & 0.941 & 0.941 & 0.955 & 0.941 & 0.970 & 0.951 & 1 & 0.906 \\
			PROPERTY  & 1 & 1 & 1 & 0.962 & 0.928 & 1 & 0.926 & 0.888 & 0.967 & 0.912 & 0.934 & 0.891 & 0.947 & 0.947 & 0.947 \\
			PROPERTY V\footnotemark[1]  & 0.987 & 1 & 0.975 & \textbf{1} & 1 & 1 & \textbf{0.941} & 0.896 & 0.991 & \textbf{1} & 1 & 1 & 0.971 & 0.943 & 1 \\
			CONDITION  & \textbf{1} & 1 &1 & \textbf{1} & 1 & 1 & 0.857 & 0.750 & 1 &\textbf{1} & 1 & 1 & \textbf{1} & 1 & 1 \\
			CONDITION V\footnotemark[1]  & 0.933 & 1 & 0.875 & \textbf{1} & 1 & 1 & 0.909 & 0.833 & 1 & \textbf{1} & 1 & 1 & \textbf{1} & 1 & 1 \\
			\botrule
	\end{tabular*}}
	
	\footnotemark[1]{PROPERTY V and CONDITION V represent PROPERTY VALUE and CONDITION VALUE, respectively.}
\end{table*}
\par
As shown in the Table \ref{tab:extraction}, F1 scores for all five entity types exceed 0.9 besides CONDITION on dataset 3, thereby establishing a robust foundation for the subsequent entity allocation task. We have highlighted the maximum F1 scores achieved for each entity type within the respective datasets. Notably, all of these peak F1 scores are attained by the PROPERTY VALUE, CONDITION and CONDITION VALUE. However, The F1 scores for the latter two types of entity exhibit greater fluctuation across different datasets. The minimum F1 scores of CONDITION and CONDITION VALUE are 0.857 and 0.909, respectively, which is significantly lower than the minimum F1 scores of other entities. This phenomenon can be attributed to the frequent omission of these two kinds of entities within sentences, leading to a limited number of golden labels. Consequently, a small number of errors can significantly impact the F1 scores. Among the first three entities with larger quantities, the model demonstrates the best performance on the PROPERTY VALUE.
This can be attributed to the fact that such entities typically manifest as numerals accompanied by units, making them more readily discernible compared to textual entities. Furthermore, the model demonstrates consistent performance on the MATERIAL entity across all test sets, with an F1 score stabilizing around 0.96. In contrast, the F1 score for the PROPERTY entity declines as the number of tuples within the sentences increases, dropping from 1 to 0.912. This decrease may be ascribed to the presence of a more heterogeneous array of PROPERTY entities, while MATERIAL entities within the same sentence exhibit a greater degree of homogeneity.  
\subsection{Performances of Entity Allocation}
The innovative aspect of the entity allocation task presented in this paper lies in its direct prediction of which entities will be assigned to the same tuple, thereby circumventing the need to predict relationships between entities. This approach not only reduces potential errors but also enhances simplicity. It is evident that the complexity of the entity allocation task varies across datasets, particularly in sentences containing a diverse number of tuples. Hence, we categorize the datasets based on the tuple count within the sentences and implement distinct training and testing procedures for each test set. 

In evaluating the performance of the model, we argue that only complete and accurate tuple can contribute to an increase in the total correct count, irrespective of the efficacy of entity extraction. Therefore, The metrics we employ to evaluate the performances of entity allocation are calculated in the following manner:
\begin{equation}
	\begin{aligned}
		\text{Precision} & = \frac{\text{Number of correct tuples extracted }}{\text{Number of tuples extracted} } \\
		\text{Recall} & = \frac{\text{Number of correct tuples extracted }}{\text{Number of tuples in test set }} \\
		\text{F1} & = \frac{2 * \text{Precision} * \text{Recall}}{\text{Precision} + \text{Recall}}
	\end{aligned}
\end{equation}

\begin{figure*}[h]
	\centering
	\includegraphics[width=1\linewidth]{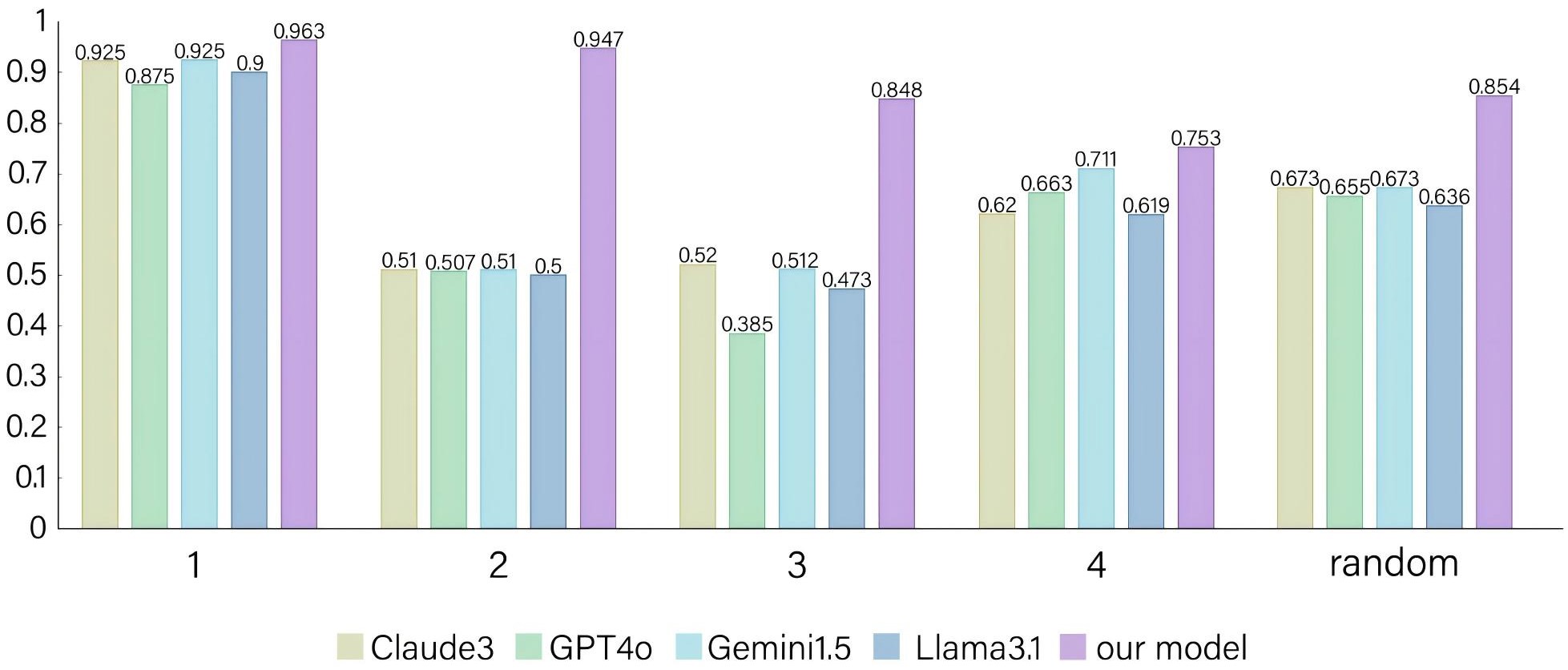}
	\caption{\textbf{The F1 scores of the purposed and baseline models}. A total of four baseline models are employed, comprising four strong large language models. Among all the models, the one we proposed achieved the best performance.}
	\label{fig:F1}
\end{figure*}

To achieve a comprehensive comparison between our proposed task-specific approach and task-agnostic large language models, we also employed a pre-training and prompting strategy utilizing four prominent models: Claude3 Haiku \cite{anthropic2024claude}, GPT4o \cite{Achiam2023GPT4TR}, Gemini1.5 \cite{reid2024gemini} and Llama3.1(70 B) \cite{dubey2024llama} as benchmarks. The overall results for both the proposed model and the reference models are presented in Table \ref{tab:allocation} (for brevity, these models are denoted as Claude3, GPT4o, Gemini1.5 and Llama3.1 in the table), while the F1 scores for all models are illustrated in Fig. \ref{fig:F1}. Furthermore, in order to To mitigate the potential issue of hallucination associated with large language models to some extent, we restrict the entity "CONDITION" to temperature within the prompt and the prompt we design is "Extract as many tuples of the form ["MATERIAL","PROPERTY","PROPERTY VALUE","TEMPERATURE","TEMPERATURE VALUE"] from the input sentences as possible. 
\begin{table*}[h]
	\caption{Performances of proposed and baseline models}\label{tab:allocation}
	\setlength{\tabcolsep}{0.2mm}{
		\begin{tabular*}{\textwidth}{@{\extracolsep\fill}l|ccc|ccc|ccc|ccc|ccc}
			\toprule%
			& \multicolumn{3}{@{}c@{}}{1} & \multicolumn{3}{@{}c@{}}{2} &
			\multicolumn{3}{@{}c@{}}{3} &
			\multicolumn{3}{@{}c@{}}{4} &
			\multicolumn{3}{@{}c@{}}{random} 
			\\\cmidrule{2-4}\cmidrule{5-16}%
			models & F1 & P & R & F1 & P & R & F1 & P & R & F1 & P & R & F1 & P & R \\
			\midrule
			Claude3  & 0.925 & 0.925 & 0.925 & 0.51 & 0.519 & 0.500 & 0.520 & 0.496 & 0.547 & 0.62 &0.861 & 0.484 & 0.673 & 0.755 & 0.607 \\
			
			GPT4o  & 0.875 & 0.875 & 0.875 & 0.507 & 0.543 & 0.475 & 0.385 & 0.493 & 0.316 & 0.663 & 0.883 & 0.531 & 0.655 & 0.734 & 0.59 \\
			
			Gemini1.5  & 0.925 & 0.925 & 0.925 & 0.510 & 0.519 & 0.500 & 0.512 & 0.474 & 0.556 & 0.711 & 0.904 & 0.586 & 0.673 & 0.755 & 0.607 \\
			
			Llama3.1  & 0.900 & 0.900 & 0.900 & 0.500 & 0.500 & 0.500 & 0.473 & 0.428 & 0.530  & 0.619 & \textbf{0.909} & 0.469 & 0.636 &0.714 & 0.574 \\

			The study  & \textbf{0.963}& \textbf{0.951} & \textbf{0.975} & \textbf{0.947} & \textbf{0.947} & \textbf{0.947} & \textbf{0.848} & \textbf{0.893} &\textbf{ 0.807} & \textbf{0.753} & {0.753} & \textbf{0.753} & \textbf{0.854} & \textbf{0.830} & \textbf{0.880} \\

			\botrule
	\end{tabular*}}
\end{table*}  
\par

As shown in the Table \ref{tab:allocation}, we have highlighted the superior metric values for each test set to facilitate the identification of the top-performing model. Our model surpasses s all other baseline models on across all test sets in terms of the F1 score.
\par
Overall, the performance of the proposed model decreases as the number of tuples contained within a single sentence (denoted as 
$k$) increases. This decline can be attributed to the increased difficulty for the model to avoid confusion during the entity matching process as more tuples are included in the input sentence. Furthermore, authors often exhibit a tendency to use abbreviations or omit certain information when sentences contain multiple tuples, resulting in a reduction of semantic information available to the model. 

Specifically, when there is only tuple ($k = 1$), the model is exclusively tasked with entity extraction, and our model achieves its optimal performance with an 
$\text{F1}$ score of 0.963. At $k=2$, the model's precision and recall are equal, as the number of tuples generated by the model matches that of the gold standard labels. Additionally, the number of entities extracted aligns with the true labels. Therefore, the results for $k=1$ and $k=2$ both underscore the efficacy of our MatSciBERT + Pointer Network approach. The slight decline in F1 to 0.947 when $k=2$ suggests the model's emerging errors due to confusion during entity allocation. 

Moreover, with each increment of $k$ by 1 starting from $k=2$, the $\text{F1}$ score achieved by our model decreases by approximately 0.1. Notably, at $k=4$, the model's performance significantly deteriorates to an $\text{F1}$ of 0.753. This decline is partly attributable to the fact that each sentence in the dataset, on average, contains four tuples for extraction, rendering the sentence structure more complex and challenging for the model. Furthermore, as the content volume increases, authors are prone to employing omissions and abbreviations that hinder the model's learning and prediction capabilities. Additionally, the limited quantity of data with sentences containing four tuples and the variability in authors' writing styles may complicate model training. 

In the last dataset (with an average $k$ of 2.17), the F1 score is 0.854, slightly higher than that for $k=3$($\Delta = +0.006$), yet markedly lower than the F1 score for $k=2$($\Delta = -0.093$). This may be due to the model struggling with randomly selected sentences that have a higher number of tuples. performing poorly on randomly selected sentences with a higher number of tuples. The significance of this dataset lies in its ability to validate the model's performance in a random context. Given that the model is considered as a correct match only when all five entities are accurately extracted and allocated, we regard an $\text{F1}$ score of 0.854 as a commendable outcome.


On the other hand, the performance of the large language models remains relatively consistent across datasets. Notably, Gemini1.5 demonstrates superior efficacy, marginally trailing Claude3 on the third dataset ($\Delta = -0.008$) while significantly outpacing Claude3 on the fourth dataset ($\Delta = +0.091$). For the remaining datasets, the F1 scores for both models are identical. At $k=1$, $\text{F1}$ scores for the large models consistently exceed 0.9, indicating their proficiency in the entity extraction task. However, in contrast to our model, the large language models exhibit diminished performance on the second and third datasets (average $\text{F1}$ = 0.507 when $k=2$, average $\text{F1}$ = 0.473 when $k=3$), although they perform better on the fourth dataset (average $\text{F1}$ = 0.653 when $k=4$), yet still fall short compared to our model.

Our analysis suggests that large models are prone to erroneously extracting excessive irrelevant content as entities, often misidentifying the concentration of a chemical element within an alloy or the methodology of metal fabrication as CONDITION entities. Additionally, these models tend to generate entities replete with extraneous information rather than extracting precise entities based on the input sentence, resulting in poorer performance. Conversely, on the fourth test set, where sentences focus more on conveying tuple-related content with minimal extraneous information, the precision of the large language models significantly improves (average P = 0.890 when $k=4$), even surpassing our model (P = 0.753 when $k=4$). Consequently, their performance on the fourth dataset is comparatively superior to that on the second and third datasets.


\subsection{Ablation Test}
Since our work uses several novel components, especially attention mechanisms, in the task of entity allocation task, we conducted a series of ablation tests to verify the importance of these components for structural information extraction in materials science literature. In particular, subsequent to the excision of several components from the model, we retrain the model and evaluate the performance on the test datasets. The components encompass the task of entity allocation, intra-entity attention and inter-entity attention (These settings are denoted as without allocation, without intra and without inter in the subsequent table). Due to the fact that the first test set does not involve entity allocation, we start the ablation tests from the second dataset. The overall result is shown in the Table \ref{tab:ablation}. We bold the best F1, precision (P), and recall (R) metrics in each test set.

\begin{table*}[h]
	\caption{The results of the ablation test}\label{tab:ablation}
	\setlength{\tabcolsep}{0.4mm}{
	\begin{tabular*}{\textwidth}{@{\extracolsep\fill}l|ccc|ccc|ccc|ccc}
		\toprule%
		& \multicolumn{3}{@{}c@{}}{our model} & \multicolumn{3}{@{}c@{}}{without allocation} &
		\multicolumn{3}{@{}c@{}}{without intra} &
		\multicolumn{3}{@{}c@{}}{without inter}
		 \\\cmidrule{2-4}\cmidrule{5-13}%
		Dataset & F1 & P & R & F1 & P & R & F1 & P & R & F1 & P & R \\
		\midrule
		Dataset 2  & \textbf{0.947} & \textbf{0.947} & \textbf{0.947} & 0.562 & 0.4 & \textbf{0.947} & 0.868 & 0.868 & 0.868 & 0.855 & 0.855 & 0.855 \\
		Dataset 3  & \textbf{0.848} & \textbf{0.893} & 0.807 & 0.448 & 0.307 & \textbf{0.833} & 0.811 & 0.854 & 0.772 & 0.793 & 0.835 & 0.754 \\
		Dataset 4  & \textbf{0.753} & \textbf{0.753} & 0.753 & 0.341 & 0.213 & \textbf{0.854} & 0.730 & 0.730 & 0.730 & 0.685 & 0.685 & 0.685 \\
		Random   & \textbf{0.854} & \textbf{0.830} & 0.880 & 0.559 & 0.405 & \textbf{0.900} & 0.835 & 0.811 & 0.860 & 0.816 & 0.792 & 0.84 \\
		\botrule
	\end{tabular*}}

\end{table*}

\begin{figure*}[h]
	\centering
	\includegraphics[width=1\linewidth]{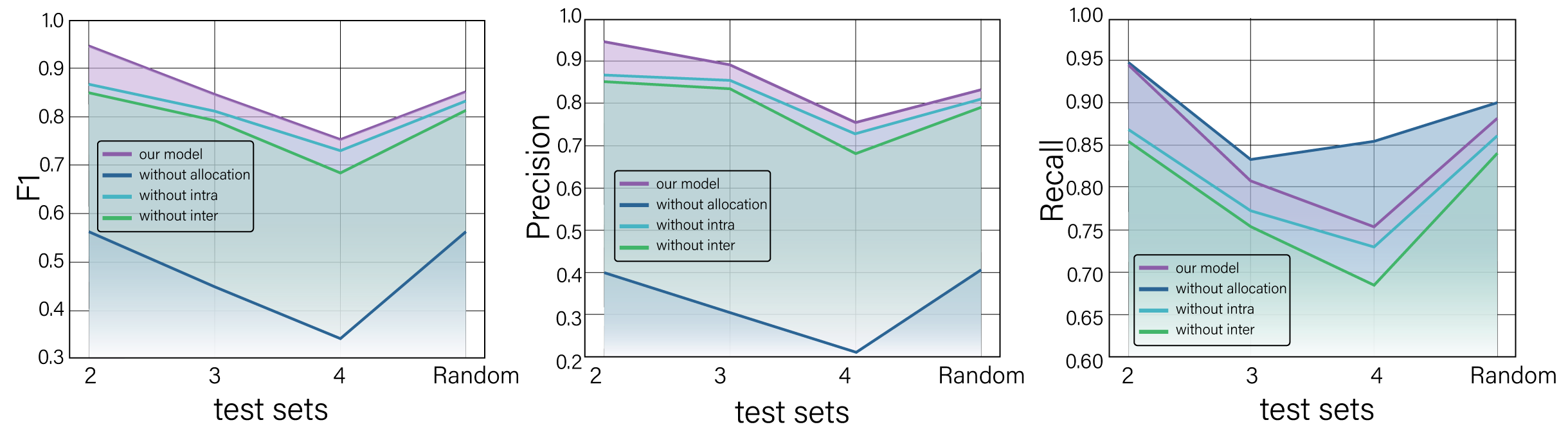}
	\caption{\textbf{The F1 scores of the purposed and variant models}. Among all the models, the one we proposed achieved the best performance.}
	\label{fig:ablation}
\end{figure*}

To enhance interpretability, the result is visualized which is shown in the Fig. \ref{fig:ablation}. Overall, the proposed model with complete components performs achieves the best F1 score on all the datasets, which indicates that the task of entity allocation, intra-entity and inter-entity attention are beneficial to the model. Regardless of the model settings, the F1 score decreases as the number of tuples to be extracted in the test set increases. Note that the model without entity allocation consistently shows the lowest F1 and precision scores, but the highest recall scores. This is because the output tuples of the model without entity allocation is all possible combinations of entities, therefore, the model exhibits poor performance in the trade-off between precision and recall, resulting in the lowest F1 score. This underscores the significance of entity allocation in our work. By the way, we also observe that the complete model also achieve the highest recall score (R = 0.947) on dataset 2, indicating that the model found all the correct tuples during the entity allocation phase without generating any incorrect ones. This suggests that the model performs best when there are only two tuples to be extracted.

Furthermore, when the entity allocation is employed, the F1 score (0.855, 0.793, 0.685, 0.816, respectively) of the model without inter-attention is consistently worse than that (0.868, 0.811, 0.730, 0.835, respectively) of the model without intra-attention which proving that inter-attention is more important than intra-attention in extracting structured information. This indicates that the model has learned the implicit relationships between different types of entities from the inter-entity attention for entity allocation. For instance, the relationship between the unit of the property values and properties, such as hardness matching with the unit Hv, and yield strength matching with the unit Pa. Intra-entity attention focuses more on the differences within representations of the same type of entity, such as positional and semantic information embedded in the representations, and it also plays a significant role.

\section{Discussion}
Our work distinguishes itself from traditional approaches in two key aspects. Firstly, we prioritize the accuracy of the structured information extracted by the model, rather than merely the efficacy of entity extraction. Secondly, instead of using the conventional named entity recognition and relation extraction (NERRE) framework, we employ an integrated methodology that combines entity extraction and matching to generate structured information. Isolated entities often lack the capacity to convey complete scientific meanings; therefore, an accurate multidimensional tuple is essential for representing the structured information within the literature. By examining multi-principal-element alloys as a case study, we meticulously define the various entities encompassed in the required tuples, placing emphasis on the efficacy of tuple extraction to enhance the practical significance and applicability of our method. Additionally, we note that it is sufficient to identify which entities belong to the same tuple without the necessity of predicting the relationships between each individual entity. When a sentence contains multiple tuples requiring extraction, our integrated entity extraction and matching approach proves to be more precise and efficient compared to the NERRE method, which is more susceptible to errors and involves unnecessary computational resource expenditure.
\par
Large language models employ a pre-training plus prompt approach, where users construct prompts and provide inputs to guide the model in performing specific tasks. In contrast, our methodology adopts a pre-training plus fine-tuning framework, which involves the deliberate design of deep learning networks and the systematic annotation of data. This approach not only enhances the model's capability but also significantly improves its performance in extracting multiple tuples, as demonstrated in this paper. Although large language models offer a more user-friendly interface, researchers encounter substantial limitations due to commercial restrictions—such as paid subscriptions and usage caps—which result in increased costs and added complexities. Researchers often find it difficult to effectively leverage output errors to refine model performance, resulting in a trade-off between convenience and accuracy. Our model, which prioritizes precision in output, addresses this issue more effectively, making it particularly suited for tasks that demand high accuracy.

Another notable drawback of large language models is the phenomenon of "hallucination"\cite{ji2023survey, farquhar2024detecting} where they generate plausible yet misleading information not based on factual data. For instance, in our context, hallucinations may result in the misrepresentation of entities, such as incorrectly extracting the MATERIAL entity "Al0.3CoCrFeMnNi" as "Y2O3-reinforced Al0.3CoCrFeMnNi." Additionally, these models can misclassify non-entity text as entities, such as interpreting the content of a specific element within an alloy as a CONDITION entity. Although some of these outputs may seem plausible, they ultimately compromise the accuracy required for successful entity extraction. Our approach, on the another hand, as a specialized tool specifically designed, offers greater precision compared to the broader but less accurate capabilities of large language models and can be easily used in other materials domains given adequate high-quality data for fine-tuning. Nonetheless, this does not imply that large language models are inherently inadequate. Firstly, the accuracy of information extraction can be enhanced through fine-tuning these models. Secondly, large language models are typically better suited for tasks that demand robust language comprehension—such as coding, reasoning, and mathematical logic—rather than more structured tasks like entity extraction, where our model demonstrates superior performance.

\par
One limitation of our approach is the requirement for high-quality data for fine-tuning, typically no less than hundreds of tuple entries. This annotation process demands significant effort and expertise and assumes the availability of a sufficient volume of relevant literature in the material domain. Our experimental results indicate that the model struggles with higher tuple extraction due to data insufficiency. A potential solution is to initially leverage large language models to screen a vast corpus of literature for potential candidates, followed by expert selection, which embodies a Human-in-the-loop annotation approach. Furthermore, when applying this model to material domains with limited data, its performance may not match that observed in fields with abundant literature. In such instances, transfer learning could be employed to enhance the model's effectiveness. Related research is currently underway.


\par
In conclusion, we propose an innovative approach integrating entity extraction and entity allocation models tailored to capture structured information in materials science literature, particularly within complex corpora containing multiple scientifically tuples. This challenge has been insufficiently addressed by previous research, and is indeed prevalent in materials science literature through our analysis. Instead of adhering to the traditional paradigm of named entity extraction followed by relation extraction, the novel strategy integrates an entity extraction model based on MatSciBERT and Pointer Network and an entity allocation model augmented by inter-entity and intra-entity attentions. Extensive experiments demonstrate that our model achieves remarkable performance, with F1 scores of 0.963, 0.947, 0.848, and 0.753 for datasets containing 1, 2, 3, and 4 tuples, respectively. Additionally, an F1 score of 0.854 was attained on a randomly selected dataset. These high F1 scores emphasize the model’s proficiency in accurately extracting structured information, showcasing its robustness across varying complexities of tuple extraction. We believe that our methodology provides a simpler, more effective, and accurate alternative to large language models in specialized tasks, empowering researchers to obtain comprehensive insights in materials science. By effectively addressing varying properties among disparate materials or the property values of a single material under different conditions, our approach aids to deliver robust, high-quality data to support data-driven strategies in materials design. 

\section{Methods}\label{sec2}
\subsection{Literature Acquisition}
The material literature used in this paper are downloaded with publisher consent from Elsevier. This work mainly focus on high entropy alloys, a large amount of paper relevant to high entropy alloys are retrieved and the top 200 papers are downloaded in HTML/XML format based on their relevance and timeliness. Then the papers are converted into text format files and form a corpus for subsequent processing. 
\subsection{Preprocessing and Annotating}
In this stage, we need to identify sentences that describe the significant properties of alloys within material papers in the corpus and annotate all tuples with correct scientific meanings within these sentences as golden labels which is time-consuming and requires expertise. There are 5 entity types in the tuples:
    \begin{itemize}
    	\item \textbf{MATERIAL} The type MATERIAL is used to annotate text spans referring to alloys. A significant portion of them are chemical formula denoting the composition of alloys (e.g., Ti–30Nb–1Mo–4Sn), and the remaining portion comprises abbreviations denoting specific materials (e.g., C.P. Ti). However, we do not view a pronoun of a specific alloy as constituting an entity of MATERIAL (e.g. that alloy or the alloy).
    	\item \textbf{PROPERTY} The type of PROPERTY is used to annotate text spans referring to names of material property. High entropy alloys are characterized by a multitude of important properties, such as elastic modulus, tensile strength and yield strength. Variability in terminology across papers may result in the same property being expressed using different terms (e.g., UTS and ultimate tensile strength). Golden label should adhere to the specific term used within the sentence.
    	\item \textbf{PROPERTY VALUE} The type of PROPERTY VALUE is used to annotate numerical values and their units of respective properties. Moreover, it includes "over", "between" and even "$>$" (e.g., over 800 Mpa and $>$ 40\%)
    	\item \textbf{CONDITION} The type of CONDITION is used to annotate specific measurement conditions associated with a given property (e.g. temperature). Its introduction is essential due to the sensitivity of certain properties to measurement conditions, and its absence would render the entire tuple incomplete from a scientific perspective. 
    	\item \textbf{CONDITION VALUE} we annotate numerical values and their
    	units of specific condition with the type CONDITION VALUE.
    \end{itemize}

Despite the frequent omission of these five entities in research papers, tuple annotation only allows for the absence of CONDITION VALUE and CONDITION VALUE. Tuples with missing information in the first three entities will not be considered golden labels. In light of the stringent requirements and laborious nature of data annotation, we extracted a subset of 568 golden tuples from 255 sentences within the corpus. To comprehensively test the performance of our model, we further divided the data into four categories based on the number of golden tuples present in the selected sentences. These categories contained 1, 2, 3, and more than 4 golden tuples, respectively. Additionally, we randomly sampled a portion of the sentences to assess the average performance.
Also, issues of garbled text, resulting from changes in the encoding format, and ambiguous expression have been rectified.
\subsection{Entity Extraction}
In our model, MatSciBERT generates vector representations of the input text for computer processing. Specifically, it is pre-trained on a large-scale corpus of materials science literature. The numerous stacked self-attention layers within the model enable it to better learn the relationships between words in the input text, thus understanding materials science-specific terminology and grammatical structures. After pre-training, it can generate vector representations of the input text. Assuming MatSciBERT generates $t$ tokens, denoted as $\textbf{t} = [t_1, …, t_n]$, the vector representations of each token are generated:
\begin{equation}
	\mathbf{x}_t = \operatorname{MatSciBERT}(t)
\end{equation}
where $t$ is one of the tokens and $\mathbf{x}_t$ is the vector representations. 
\par
After obtaining the vector representations, pointer network trains two binary classifiers to predict the positions of the start and end tokens of the desired entity within the input text. Once the start and end tokens of the entity are identified, the entity can be extracted. In our model, it calculates the probability of each token generated by MatSciBERT being the start and end token of the desired entity, fully connected layers and activation function are used to calculate the probabilities of the head pointer and tail pointer:
\begin{equation}
	\begin{aligned}
		& P_s^{\text {material }}(t)=\operatorname{Sigmoid}\left(\mathbf{W}_s^{\text {material }} \cdot \mathbf{x}_t+\mathbf{b}_s\right) \\
		& P_e^{\text {material }}(t)=\operatorname{Sigmoid}\left(\mathbf{W}_e^{\text {material }} \cdot \mathbf{x}_t+\mathbf{b}_e\right)
	\end{aligned}
\end{equation}
where $P_s^{\text {material }}(t)$ and $P_e^{\text {material }}(t)$ are probabilities of $t$ to be the start and end token of MATERIAL entity, $\mathbf{W}_s^{\text {material}}$, $\mathbf{b}_s$, $\mathbf{W}_e^{\text {material }}$ and $\mathbf{b}_e$ are matrices of learnable parameters specific for head pointer and tail pointer, respectively. Furthermore, if we want to calculate the probabilities for other entity types, new learnable parameters must be set. In summary, the pointer network essentially trains five sets of binary classifiers to determine whether a token is a head or tail pointer. 
\par
After probability calculations are complete, tokens with probabilities exceeding a predefined threshold, which is a manually set hyperparameter, are considered pointers. Therefore, a list containing only \textbf{0}s and \textbf{1}s are generated. Each position in the list represents whether the corresponding token is a pointer (\textbf{1}) or not (\textbf{0}):
\begin{equation}
	\begin{aligned}
		\hat{L}_s^{\text {material }}\left(t_i\right) & =\left\{\begin{array}{l}
			1, P_s^{\text {material }}(t) \geq \beta_s^{\text {material }} \\
			0, P_s^{\text {material }}(t)<\beta_s^{\text {material }}
		\end{array}\right. \\
		\hat{L}_e^{\text {material }}\left(t_i\right) & =\left\{\begin{array}{l}
			1, P_e^{\text {material }}(t) \geq \beta_e^{\text {material }} \\
			0, P_e^{\text {material }}(t)<\beta_e^{\text {material }}
		\end{array}\right.
	\end{aligned}
\end{equation}
where $\beta_s^{\text {material }}$ and $\beta_e^{\text {material }}$ are the thresholds specific for MATERIAL entity.
\par
As for training, our objective is to minimize the following  cross-entropy loss function:
\begin{equation}
\begin{aligned}
\mathcal{L}_{1}= -\dfrac{1}{|\mathcal{T}||\textbf{t}|}
	\sum_{\tau \in \mathcal{T}}\sum_{t=1}^{|\textbf{t}|}
	[L_{s}^{\tau}(t) \log (P_s^{\tau}(t)) + \\
 L_{e}^{\tau}(t) \log (P_e^{\tau}(t)) ]
\end{aligned}
\end{equation}
where $\mathcal{T}$ is the set of entity types, $\textbf{t}$ is the set of tokens and $L_{s}^{\tau}(t)$ and $L_{e}^{\tau}(t)$ are golden labels indicating where $t$ is a head or tail pointer. While testing, text spans referring to various types of entity can be obtained using a simple heuristic method. In particular, the head pointer and the nearest tail pointer after it form an entity. 

\subsection{Entity Allocation}
\label{subsection:allocation}
We established the entity matching score matrix, which allows the generation of all possible tuple matching patterns and enables supervised training based on labels (as shown in the Fig. \ref{fig:framework} (\textbf{a}) and (\textbf{b})). Assuming there are only two types of entities that could be confused, as shown in the Fig. \ref{fig:framework}(\textbf{b}), and there are $n$ and $m$ entities of the two types respectively. Their vector representations are denoted as $[h_1,h_2,...,h_n]$ and $[g_1,g_2,...,g_m]$ (the representations of an entity is obtained by summing the vector representations of all its tokens). When calculating the inter-attention representations, we first need to determine the semantic correlation between the two entity representations. Take $h_i $ and $g_j$ as an example, we have:
\begin{equation}
	S_{i j}=\frac{1}{\sqrt{d}} \sigma\left(h_i, g_j\right)
\end{equation}
where $S_{i j}$ is the semantic correlation between $h_i $ and $g_j$, where $ \sigma$ denotes dot product operator, $d$ denotes
embedding dimension of $h_i $ and $g_j$ \cite{vaswani2017attention}.
\par
Based on  $S_{i j}$, we apply inter-attention mechanism to generate get a $g_j$ aware representations for $h_i$ and a $h_i$ aware representations for $g_j$:
\begin{equation}
	\begin{aligned}
		\boldsymbol{A}_i^{g 2 h} & =\sum_{j=1}^m S_{i j} \cdot g_j \\
		\boldsymbol{A}_{{j}}^{h 2 g} & =\sum_{i=1}^n S_{i j} \cdot h_i
	\end{aligned}
\end{equation}
where $\boldsymbol{A}_i^{g 2 h}$ and $\boldsymbol{A}_{{j}}^{h 2 g}$ are representations of $h_i$ and $g_j$, respectively. They are also represented by the vectors in the upper box of Fig. \ref{fig:framework} (\textbf{c}).
\par
Intra-entity attention allows the model to learn implicit relationships between entities of the same type. For example, among PROPERTY VALUE entities, the size or unit relationships between the values can help to avoid confusion between PROPERTY VALUE and PROPERTY. The calculation methods for the two attention mechanisms are similar, for $h_i$ we have:
\begin{equation}
	\begin{gathered}
		\mu_{i j}=\operatorname{Softmax}\left(\frac{1}{\sqrt{d}} \sigma\left(h_i, h_j\right)\right) \\
		\boldsymbol{A}_i^{h 2 h}=\sum_{j=1}^n \mu_{i j} \cdot h_j
	\end{gathered}
\end{equation}
where $\mu_{i j}$ is the semantic correlation between $h_i $ and $h_j$, where $ \sigma$ denotes dot product operator, $d$ denotes embedding dimension of $h_i $, $\boldsymbol{A}_i^{h 2 h}$ is the representation of $h_i$.  For $g_j$ we have:
\begin{equation}
	\begin{gathered}
		v_{j k}=\operatorname{Softmax}\left(\frac{1}{\sqrt{d}} \sigma\left(g_j, g_k\right)\right) \\
		\boldsymbol{A}_j^{g 2 g}=\sum_{k=1}^m v_{j k} \cdot g_k
	\end{gathered}
\end{equation}
where $v_{i j}$ is the semantic correlation between $g_j $ and $g_k$ and $\boldsymbol{A}_i^{g 2 g}$ is the representation of $g_j$.
\par
Finally, for the two entities that may be confused, we concatenate the six obtained vector representations and transform the entity allocation task into a binary classification problem:
\begin{equation}
	\hat{z}_{i j}=\mathbf{U}_{h g}\left(\left[h_i ; g_j ; \boldsymbol{A}_i^{g 2 h} ; \boldsymbol{A}_{\mathrm{j}}^{h 2 g} ; \boldsymbol{A}_i^{h 2 h} ; \boldsymbol{A}_j^{g 2 g}\right]\right)
\end{equation}
where $\hat{z}_{i j}$ is the matching score for $h_i $ and $g_j$ and $U_{h g}$ is a learnable parameter. 
\par
As for training, our objective is to minimize the following cross-entropy loss function:
\begin{equation}
	\mathcal{L}_{2}= -\dfrac{1}{|\mathcal{H}||\mathcal{G}|}
	\sum_{h \in \mathcal{H}}\sum_{g \in \mathcal{G}}
	{z}_{i j} \log (\hat{z}_{i j}) 
\end{equation}
where $\mathcal{H}$ and $\mathcal{G}$ are sets of two types of entity, ${z}_{i j}$ is the golden label indicating whether $h_i$ and $g_j$ are in the same tuple. During testing, $h_i$ will choose $g_j (j = 1,2,...,m)$ that maximizes the matching score $\hat{z}_{ij}$. If the number of entity types that might be confused is $x (x > 2)$, the model will match each PROPERTY VALUE entity with other four kinds of entities until all the PROPERTY VALUE entities are assigned into complete tuples. During test, the fully connected layer and activation function serve to transform the $\hat{z}_{ij}$ into probabilities that $h_i$ and $g_j$ are the correct matching. After calculating all the $z_{ij}$ in the matching matrix as probabilities, the correct matches are given by the $h_i$ and $g_j$ corresponding to the maximum probability in each row.
\par
Moreover, our research reveals that the two types of entities that may be confused often occur in equal numbers within a sentence. In such instances, their correct alignment typically follows their sequential occurrence in the text. For example, in the sentence, "\textit{This new HfNbTiZr alloy has the tensile YS of 879 MPa, UTS of 969 MPa, and the plastic elongation of 14.9\%}", the sequence of PROPERTY is ["YS", "UTS", "plastic elongation"], which corresponds to the sequence of PROPERTY VALUE entities ["879 MPa", "969 MPa", "14.9\%"]. By matching these entities in order, one achieves the correct pairings: "YS" with "879 MPa", "UTS" with "969 MPa", and "plastic elongation" with "14.9\%". These associations are accurately represented in the matching matrix, where the diagonal elements indicate the correct alignments. Consequently, during the testing and inference phases, we introduce a parameter $\lambda$ to the diagonal of the matching matrix, thereby augmenting the matrix with rule-based reinforcement.

%
%

\section*{Data availability}
Any data used for the work is available from the corresponding authors upon reasonable request. The JSON-formatted dataset used in this work is available at \href{https://github.com/HEI-MENGZHE/Material\_Tuple/tree/main/data}{https://github.com/HEI-MENGZHE/Material\_Tuple/tree/main/data}

\section*{Code availability}
All the codes used in the present work are available at \href{https://github.com/HEI-MENGZHE/Material\_Tuple}{https://github.com/HEI-MENGZHE/Material\_Tuple}
\section*{Declarations}
The authors declare no competing interests.
\section*{Author contribution}
M Hei: Data Curation, Model Construction, Writing-Original Draft. Z Zhang: Analysis, Methodology, Writing-Review and Editing. Q Liu: Supervision. Y Pan: Data Curation and Funding Acquisition. X Zhao: Supervision. Y Peng: Data Curation and Visualization. Y Ye: Project Administration. X Zhang: Supervision and Methodology. S Bai: Supervision and Funding Acquisition. All authors have reviewed the manuscript.
\section*{Acknowledgement}
This work is supported by National Natural Science Foundation of China under the grants of NO.62102431 and No.U20A20231 and National University of Defense Technology Independent Scientific Innovation Fund Project.
\noindent
\bigskip
%
%
%
%

\begin{appendices}




\end{appendices}


\bibliography{paper.bib}

\end{document}